\documentclass[a4paper]{article}

\usepackage{INTERSPEECH2022}
\usepackage{booktabs}
\usepackage{siunitx}
\usepackage{microtype}

\usepackage{amssymb}
\usepackage{pifont}

\title{Tokenwise Contrastive Pretraining for Finer Speech-to-BERT Alignment in End-to-End Speech-to-Intent Systems}
\name{Vishal Sunder$^1$, Eric Fosler-Lussier$^1$, Samuel Thomas$^2$, Hong-Kwang J. Kuo$^2$, Brian Kingsbury$^2$}
\address{
  $^1$The Ohio State University, Columbus, OH, USA \\
  $^2$IBM Research AI, Yorktown Heights, NY, USA}
\email{sunder.9@osu.edu, fosler@cse.ohio-state.edu, \{sthomas, hkuo, bedk\}@us.ibm.com}

\begin{document}

\maketitle
\begin{abstract}

Recent advances in End-to-End (E2E) Spoken Language Understanding (SLU) have been primarily due to effective pretraining of speech representations. One such pretraining paradigm is the distillation of semantic knowledge from state-of-the-art text-based models like BERT to speech encoder neural networks. This work is a step towards doing the same in a much more efficient and fine-grained manner where we align speech embeddings and BERT embeddings on a token-by-token basis. We introduce a simple yet novel technique that uses a cross-modal attention mechanism to extract token-level contextual embeddings from a speech encoder such that these can be directly compared and aligned with BERT based contextual embeddings. This alignment is performed using a novel tokenwise contrastive loss. Fine-tuning such a pretrained model to perform intent recognition using speech directly yields state-of-the-art performance on two widely used SLU datasets. Our model improves further when fine-tuned with additional regularization using SpecAugment especially when speech is noisy, giving an absolute improvement as high as 8\% over previous results.
\end{abstract}

\section{Introduction}


Traditionally, spoken language understanding (SLU) is a two-step process. In the first step, speech is converted to text by an upstream automatic speech recognizer (ASR), and in the second step, a downstream natural language understanding (NLU) component is used to extract semantic labels from the recognized text. Historically, ASR and NLU were treated as two separate problems which has led to progress in both fields independently, with much less attention paid to joint studies \cite{faruqui2021revisiting}.

While a lot of progress has been made in ASR and NLU research, the cascaded nature of traditional SLU systems has posed two main problems. First, state-of-the-art (SOTA) ASR (\cite{hsu2021hubert, baevski2020wav2vec}) and NLU (\cite{liu2019roberta, brown2020language}) models are very large with hundreds of millions of parameters. To utilize them for real-world SLU problems, a cascading of these large models is inevitable, which leads to an even larger model. This makes them impractical for building real-world voice assistants that need to be fast and may need an on-device deployment.

The second problem with cascaded ASR-NLU is that errors from upstream ASR can prove to be catastrophic for the NLU component. This has an adverse effect on SLU performance \cite{gopalakrishnan2020neural, kim2021robust}. Many techniques have been proposed recently to deal with ASR errors in SLU systems \cite{stiff2019improving, serai2022hallucination, namazifar2021warped}, but how to do it most effectively still remains an open question.

Due to the above challenges with cascaded ASR-NLU systems, end-to-end (E2E) SLU has gained popularity in the recent past. Unlike their cascaded counterpart, E2E systems are extremely compact making their deployment very simple and they also alleviate the effects of ASR errors to a large extent by operating on speech directly, bypassing the use of ASR transcripts. 

To build E2E SLU systems, a key ingredient is the pre-training of neural speech encoders for learning robust speech representations. The speech encoders are then fine-tuned with a relevant criterion for downstream SLU applications.

\subsection{Related Work}
Various forms of pretraining methods have been proposed in recent work. Broadly, they can be divided into three categories: sequence-to-sequence (seq2seq) pretraining, cross-modal embedding alignment and BERT-like pretraining \cite{devlin2019bert}.

Seq2seq pretraining is an ASR based pretraining, the simplest form of which is to train an ASR model on a large out-of-domain dataset and to fine-tune the speech encoder from the trained ASR model for downstream SLU tasks \cite{lugosch2019speech, ghannay2018end}. An improved version of this is to use SLU labels when available in the pretraining step by learning to decode the transcript combined with the SLU labels \cite{kuo2020end}. This makes the model suitable for downstream SLU tasks. Cross-modal embedding alignment involves explicitly minimizing the distance between speech embeddings and the text embeddings from state-of-the-art text encoders like BERT \cite{huang2020leveraging,kim2021two}. Thus, the speech embeddings that are used for downstream tasks are made to share a common embedding space with the textual embeddings leading to better performance. Finally, BERT-like pretraining methods are inspired from transformer pretraining in SOTA NLU systems like masked language modelling (MLM) and next sentence prediction (NSP) \cite{hsu2021hubert, baevski2020wav2vec, kim2021st}. These techniques are adapted for speech processing to get speech-based pretrained transformer models.

Very often, a combination of the above pretraining categories has been shown to perform well for SLU. In particular, Rongali et al.  \cite{rongali2021exploring} use an ASR-SLU seq2seq pretraining combined with the MLM criterion on speech units obtained from forced-alignments. A combination of speech-based MLM along with an explicit cross-modal alignment between speech and text embeddings was used by Chung et al. \cite{chung2021splat} as a pretraining step. Qian et al. \cite{qian2021speech} trained an auto-regressive transformer to predict masked text tokens conditioned on speech. 

\subsection{Our Contribution}
In this work, we propose a novel pretraining method for aligning BERT-based text embeddings with speech embeddings. Hence, our work falls in the second category of cross-modal embedding alignment. But unlike previous work, our proposed methodology performs a fine-grained alignment at the token level between BERT and speech embeddings without any supervision. Most previous work in this category align a sequence-level, pooled representation of speech and its corresponding text which is typically the [CLS] token representation of BERT. Although, Chung et al. \cite{chung2021splat} have proposed a token-level alignment strategy, they have also shown that it performs worse than a simple sequence-level alignment. The motivation for our work is that there remains an untapped potential to extract more knowledge from BERT through its token-level representation which can be useful for SLU tasks. There has been limited prior work in this regard and our work is a step towards achieving BERT-like performance from speech-based models by learning finer embedding alignments between the two modalities.

In particular, our pretraining strategy follows a simple idea where we utilize the representation of a spoken utterance from a speech encoder to convert non-contextual word embeddings of the corresponding transcript to contextual word embeddings by using a cross-modal attention mechanism. The contextual word embeddings are then aligned with the embeddings from BERT of the same transcript on a token-by-token basis via a novel use of the contrastive loss \cite{chen2020simple}. This mechanism implicitly injects fine-grained semantic knowledge from BERT into the speech representation. The proposed training procedure is agnostic to the underlying speech encoder architecture and can be used with any SOTA speech encoders. We pretrain our model on 960 hours of speech-text paired Librispeech data using this technique and directly fine-tune it for the downstream speech-to-intent (S2I) task without needing any gold transcripts for the S2I data. 

Using this strategy, we achieve SOTA results on the SNIPS-Smartlights SLU dataset for the intent recognition task on both close-field and far-field subsets. We get further improvements when we apply SpecAugment data augmentation during model fine-tuning. Our model also performs on par with other SOTA models on the Fluent Speech Commands (FSC) dataset in both full resource and low resource settings.

\section{Proposed Methodology}

\subsection{Speech Encoder Architecture}
For the speech features, we use 80-dimensional log-Mel filterbank features (LFB) over 25ms frames every 10ms from the input speech signal. These are global mean-variance normalized before being fed to the speech encoder.

 The speech encoder is a 9-layer bidirectional LSTM (BiLSTM) with a single self-attention layer on top. The first 3 layers of the BiLSTM have a pyramid structure \cite{chan2016listen} which reduces the frame rate by a factor of 8. This reduction of frame rate is important both from a computational point of view as well as a key factor in learning meaningful cross-modal attention weights as shown by Chan et al. \cite{chan2016listen}. The output of the pyramid BiLSTM is transformed using a linear layer to match the dimensions of the BERT embeddings. Each BiLSTM layer also includes a residual connection followed by layer-normalization following T{\"u}ske et al. \cite{tuske2020single}. After the 9th BiLSTM layer, we add a dot-product self-attention layer with 12 attention heads following Vaswani et al. \cite{vaswani2017attention}. Between consecutive layers of the speech model we also add 10\% dropout for regularization.
 
\subsection{Tokenwise Contrastive Pretraining}

\begin{figure}[h]
    \hfill
    \centering
    \centerline{\includegraphics[width=\columnwidth]{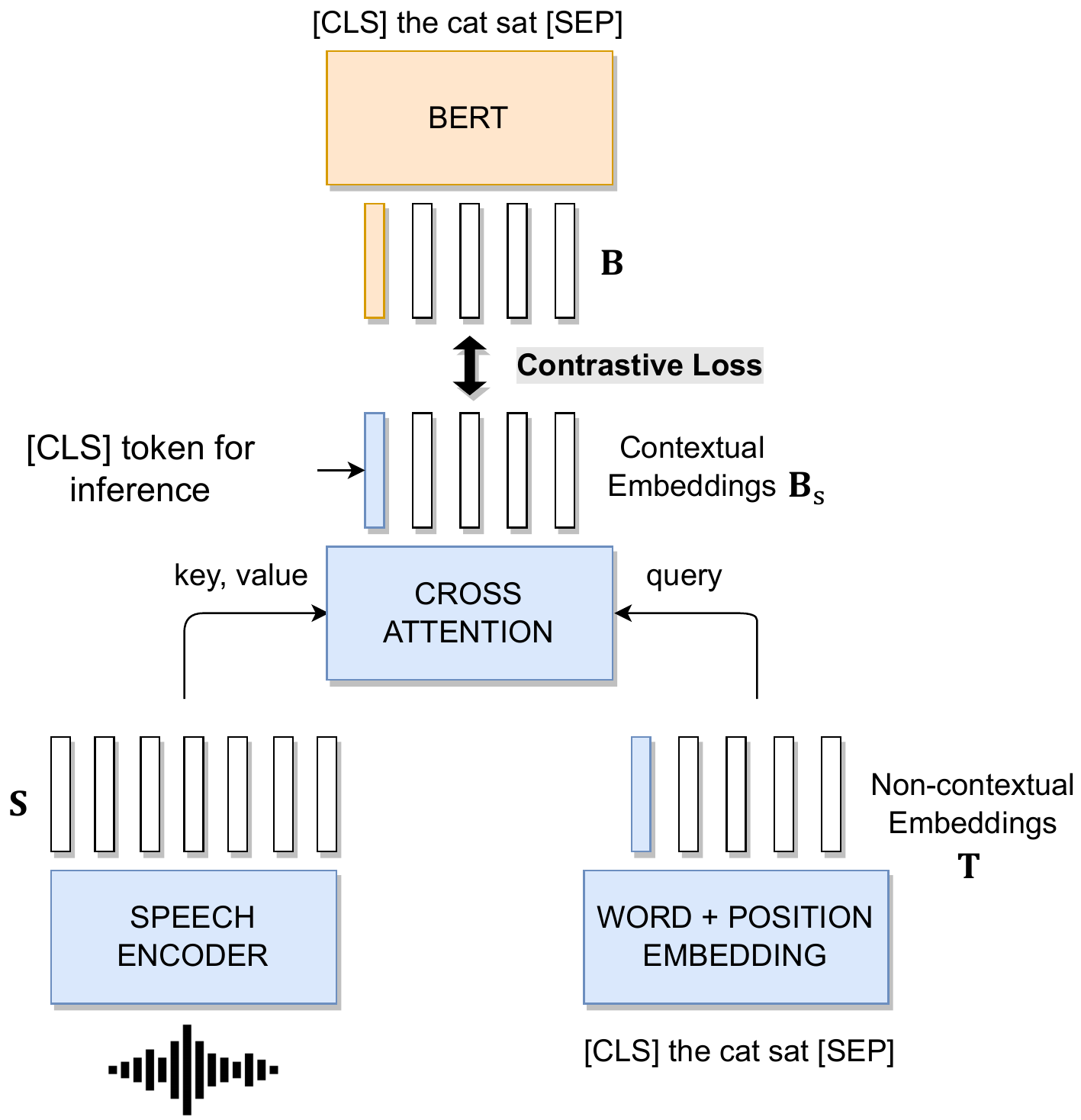}}
\caption{Model overview. During pretraining, the cross-attention mechanism yields contextual embeddings for the all input tokens using corresponding speech. During fine-tuning for downstream S2I, when only speech is available, only the [CLS] token is required.}
\label{fig:model_over}
\end{figure}

An overview of the proposed framework is shown in Figure \ref{fig:model_over}. Let $U_S$ denote a spoken utterance and $U_T$ be its transcription. The speech encoder takes $U_S$ as input and returns a representation denoted by a matrix $\textbf{S} \in \mathbb{R}^{n \times 768}$, where $n$ is the number of speech frames. A non-contextual (NC) representation of $U_T$ is obtained from a randomly initialized word embedding\footnote{It is possible to use different pretrained embeddings for the NC representation but we leave their exploration for future work.} which takes a sequence of WordPiece tokens of $U_T$ prepended and appended by the [CLS] and [SEP] tokens, repsectively. As these embeddings are non-contextual, it is important to distinguish between identical tokens at different positions. Hence, we add absolute positional encodings to the output of the NC word embedding. The NC representation of $U_T$ is denoted by a matrix $\textbf{T} \in \mathbb{R}^{m \times 768}$, where $m$ is the number of WordPiece tokens.

We use a pretrained BERT model\footnote{\texttt{https://huggingface.co/bert-base-uncased}} as a teacher for our speech encoder which is kept frozen. $U_T$ is fed to BERT and the output is denoted as $\textbf{B} \in \mathbb{R}^{m \times 768}$.\\

\noindent \textbf{Cross-Modal Attention:} We use the speech representation $\textbf{S}$ to inject contextual information into the NC embeddings $\textbf{T}$ such that the resulting contextual embeddings implicitly derive context from speech. To do this, we employ a cross-modal attention mechanism which we describe below.

The cross-modal attention follows a query-key-value based mechanism \cite{vaswani2017attention}. The NC embeddings $\textbf{T}$ act as the query and the speech embeddings $\textbf{S}$ act as the keys and values. The corresponding representations $\textbf{Q} \in \mathbb{R}^{m \times 768}$, $\textbf{K} \in \mathbb{R}^{n \times 768}$ and $\textbf{V} \in \mathbb{R}^{n \times 768}$ for the same are obtained as,
\begin{equation*}
    \begin{split}
        \textbf{Q} = \textbf{T}\textbf{W}_q \\
        \textbf{K} = \textbf{S}\textbf{W}_k \\
        \textbf{V} = \textbf{S}\textbf{W}_v \\
    \end{split}
\end{equation*}
where $\textbf{W}_q$, $\textbf{W}_k$ and $\textbf{W}_v$ $\in \mathbb{R}^{768\times 768}$ are learnable weights. The contextual embeddings $\textbf{B}_s \in \mathbb{R}^{m \times 768}$ are now obtained as,
\begin{equation*}
    \begin{split}
        \textbf{B}_s = \texttt{softmax}(\textbf{QK}^\mathbb{T})\textbf{V} \\
    \end{split}
\end{equation*}
Thus, the non-contextual word embeddings $\textbf{T}$ are converted to contextual word embeddings $\textbf{B}_s$.\\

\noindent \textbf{Contrastive Loss:} The contextual representation $\textbf{B}_s$ can now be aligned with the semantically rich BERT contextual representation $\textbf{B}$ on a token-by-token basis as they have the same sequence length $m$. For this, we employ a contrastive loss between pairs of token representations. 

All the output sequences in a batch of size $|B|$ are row-wise concatenated such that $\textbf{B}$ and $\textbf{B}_s$ are now $\in \mathbb{R}^{b \times 768}$ where $b$ is the sum of all sequence lengths in a batch ($b = \sum_{i=1}^{|B|} m_i$).

The cosine similarity between rows $i$ and $j$ in $\textbf{B}$ and $\textbf{B}_s$ is defined as,
\begin{equation*}
    \begin{split}
        s_{ij} = \textbf{B}_i\textbf{B}_{sj}^\mathbb{T}/(\tau\lVert\textbf{B}_i\rVert\lVert\textbf{B}_{sj}\rVert)
    \end{split}
\end{equation*}
where $\tau$ is a temperature hyperparameter. Then, the contrastive loss is defined as,
\begin{equation*}
    \begin{split}
        \mathcal{L}_{con} = -\frac{\tau}{2b}\sum_{i=1}^{b}(&\log\frac{\exp(s_{ii})}{\sum_{j=1}^{b} \exp(s_{ij})} + \log\frac{\exp(s_{ii})}{\sum_{j=1}^{b}\exp(s_{ji})})
    \end{split}
\end{equation*}
This is a tokenwise contrastive loss which, when optimized, brings the representations of the same tokens (positive pairs) from two modalities close together and pushes apart different tokens (negative pairs). With a larger batch size, the number of negative pairs increases, leading to a more compact embedding space. Note that in our formulation, even with a relatively small batch size, the number of negative pairs can be much larger as we perform a tokenwise comparison, i.e. \mbox{$b = (\sum_{i=1}^{|B|} m_i) \ge |B|$} as sequence length $m_i$ is always at least one.

We use speech-text pairs from 960 hours of Librispeech data \cite{panayotov2015librispeech} for pretraining. The model was trained on a single Quadro P6000 GPU for 600k steps using a batch size of 64 utterances and the AdamW optimizer with a learning rate of 1e-4. The temperature hyperparameter $\tau$ was set to 0.07.

\subsection{Fine-tuning}
Once we train a neural model using the above tokenwise contrastive pretraining method, we fine-tune it for the downstream S2I task. We assume a realistic situation where no transcripts are available for the S2I dataset. Thus, only the learnt NC embedding for the [CLS] token is used to attend over the speech encoder output through the cross attention layer. This gives a contextual BERT-like [CLS] token representation as shown in Figure \ref{fig:model_over} which is then used for inference.

The [CLS] token representation is passed through a single linear layer for classification. The entire model is fine-tuned E2E with a learning rate of 2e-5 using the AdamW optimizer.\\

\noindent \textbf{Data Augmentation:} Noisy speech can be catastrophic for any E2E SLU system, hence it is important to regularize the neural network with proper data augmentation techniques \cite{kim2021two}. To this end, we utilize the widely used SpecAugment \cite{park2019specaugment} technique during fine-tuning. We disable time-warping and only use frequency and time masking with mask widths of 15 and 70 respectively (The SM policy used in Park et al. \cite{park2019specaugment}).

\section{Experiments}

\subsection{Datasets}

For pretraining, we used 960 hours of Librispeech data. For downstream tasks, we used two popular S2I datasets: SNIPS-Smartlights and Fluent Speech Commands.\\

\noindent \textbf{SNIPS-Smartlights \cite{saade2019spoken}} is a dataset of spoken commands to a smart lights assistant. It has 1,660 utterances from 52 speakers classified into 6 unique intents. The speech in this dataset has close-field and far-field variants to test against varying degrees of noise in the environment. Following previous work, we perform a 10-fold cross-validation on this dataset.\\

\noindent \textbf{Fluent Speech Commands (FSC) \cite{lugosch2019speech}} is also a dataset of spoken commands to a home assistant. It has 23,132 training utterances from 77 speakers, 3,118 validation utterances from 10 speakers and 3,793 utterances from 10 speakers in the test set. There are 31 unique intents. Following previous work, we also train the model on a low-resource version of the dataset with only 10\% data in the training set. We create 10 disjoint subsets of the low-resource version and report the average performance.

\subsection{Results and Discussion}

\begin{table}
    \centering
    \caption{10-fold cross-validation on SNIPS-Smartlights and test accuracies on FSC dataset. The last 4 rows are variants of our model. $*$ means that the model is not comparable with others as it does not perform cross-validation but uses a fixed test set.}
    \resizebox{\columnwidth}{!}{
    \begin{tabular}{@{}lccccc@{}}\toprule
        & \multicolumn{2}{c}{SNIPS-Smartlights} && \multicolumn{2}{c}{FSC} \\
        \cmidrule{2-3} \cmidrule{5-6}
        Model & Close-field & Far-field && Full & 10\% \\
        \midrule
        \textit{S2I Transcripts used}\\
        \midrule
        \hspace{3mm}Cha et al. \cite{cha2021speak} & 80.12 & - && 99.2 & - \\
        \hspace{3mm}Rongali et al. \cite{rongali2021exploring} & 84.88 & 74.64 && - & - \\
        \hspace{3mm}ST-BERT \cite{kim2021st} & 86.91 & 69.40 && 99.6 & 99.25 \\
        \hspace{3mm}Kim et al. \cite{kim2021two} & 95.5$^*$ & 75.0$^*$ && 99.7 & 99.5 \\
        \midrule
        \textit{S2I transcripts not used}\\
        \midrule
        \hspace{3mm}Lugosch et al. \cite{lugosch2019speech} & - & - && 98.8 & 97.9 \\
        \hspace{3mm}Rongali et al. \cite{rongali2021exploring} & 81.87 & 67.83 && - & - \\
        \hspace{3mm}ST-BERT \cite{kim2021st} & 84.65 & 60.98 && 99.5 & 99.13 \\
        \hspace{3mm}Kim et al. \cite{kim2021two} & 81.3$^*$ & 51.2$^*$ && 99.0 & 98.5 \\
        \hspace{3mm}Chung et al. \cite{chung2021splat} & - & - && 99.5 & - \\ 
        \midrule
        \hspace{3mm}Sequence Contrastive & 80.90 & 62.59 && 98.9 & 97.5 \\
        \hspace{6mm}+SpecAug & 83.10 & 71.62 && 99.4 & 98.2 \\
        \hspace{3mm}Tokenwise Contrastive & 90.42 & 72.77 && 99.5 & 99.0 \\
        \hspace{6mm}+SpecAug & \textbf{92.23} & \textbf{83.07} && \textbf{99.6} & \textbf{99.4} \\
        \bottomrule
    \end{tabular}}
    \label{tab:results}
\end{table}

The results of our experiments on SNIPS-Smartlights and FSC are shown in Table \ref{tab:results}. We compare our proposed method with several other techniques proposed in recent literature. Note that all these techniques use 960 hours of Librispeech data to do some form of pretraining. This makes our comparisons fair.

The bottom four rows of Table \ref{tab:results} are variants of our proposed model which also serve as an ablation study. \textit{Sequence Contrastive} is a variant of our contrastive learning framework where we do not perform contrastive learning at the token level but rather use a pooled representation from the speech encoder as the [CLS] token similar to Agrawal et al. \cite{agrawal2020tie}. This token is then aligned with the [CLS] token from BERT using the contrastive loss. \textit{Tokenwise Contrastive} is our proposed technique. \textit{+SpecAug} means adding SpecAugment during fine-tuning.

We divide various techniques in Table \ref{tab:results} into two parts. The first part uses the in-domain S2I transcripts in some form to adapt the models to in-domain speech. The second part is a more realistic scenario where in-domain S2I transcripts are not available to perform the adaptation step. In these cases we just fine-tune the unadapted pretrained model to perform SLU directly. This tests the models on their generalization capabilities.

On the close-field subset of SNIPS-Smartlights dataset, the proposed pretraining outperforms all baselines by a significant margin. Our model beats the SOTA baseline on this dataset by 3.5\% absolute. It is worth noting that our models do not use any in-domain transcripts but still outperform baselines that use the S2I transcripts for adaptation. This demonstrates the strong generalization capability of our model. By adding SpecAugment during fine-tuning, we see a further improvement in the performance. On the far-field subset this improvement is even larger, 10.3\% absolute, which is even better than some close-field baselines. We hypothesize that this is because SNIPS-Smartlights is a low-resource dataset and for its far-field subset, it is also noisy. Therefore, SpecAugment acts as a very good regularizer that deals with both these issues simultaneously.

On the FSC dataset, we perform experiments in full-resource and low-resource settings where we only use 10\% of the training data. Without SpecAugment and not using any in-domain transcripts, our model performs on par with the SOTA ST-BERT baseline without transcripts. When we add SpecAugment, we see an improvement in both full-resource and low-resource settings. This improvement is more significant in the low-resource settings which shows that our model is capable of utilizing SpecAugment in an efficient manner. Compared with Kim et al. \cite{kim2021two}, our model is only slightly behind, 0.1\% absolute. We argue that as the performances on FSC are close to 100\%, such a small difference may not be significant. Besides, unlike Kim et al. \cite{kim2021two}, we do not use any S2I transcripts.

Note that it is not straightforward to add data augmentation techniques like SpecAugment to the previously SOTA model, ST-BERT, because this model uses phonemes from a frozen acoustic model as input rather than speech directly. Our model is truly E2E in that sense and all data augmentation techniques that can be used in ASR systems apply directly to our model.

Previously proposed ASR-based seq2seq pretraining techniques \cite{lugosch2019speech, rongali2021exploring} can also be seen as capturing some token level information but still fall short compared to the proposed method. We hypothesize that because our method performs tokenwise alignment directly in BERT's embedding space, the knowledge in our pretrained model is already semantically richer. An ASR based encoder can map speech to tokens but it is highly unlikely that the resulting embeddings would lie in BERT's space.

It is worth mentioning that compared with previous work, our neural network is very compact with only 48 million parameters. Such a compact E2E S2I model can be very useful from an on-device deployment point of view. Most previous work in Table \ref{tab:results} are transformer based \cite{kim2021two, kim2021st, rongali2021exploring, chung2021splat} and contain a lot more parameters. We kept the speech encoder as simple as possible such that most of the improvement comes from the proposed training methodology rather than the neural architecture. That said, the proposed framework is model agnostic and can be used with much larger models like HuBERT \cite{hsu2021hubert} or wav2vec2.0 \cite{baevski2020wav2vec} which may lead to better performances. This can be explored in future work.\\

\begin{figure}[h]
    \hfill
    \centering
    \centerline{\includegraphics[width=\columnwidth]{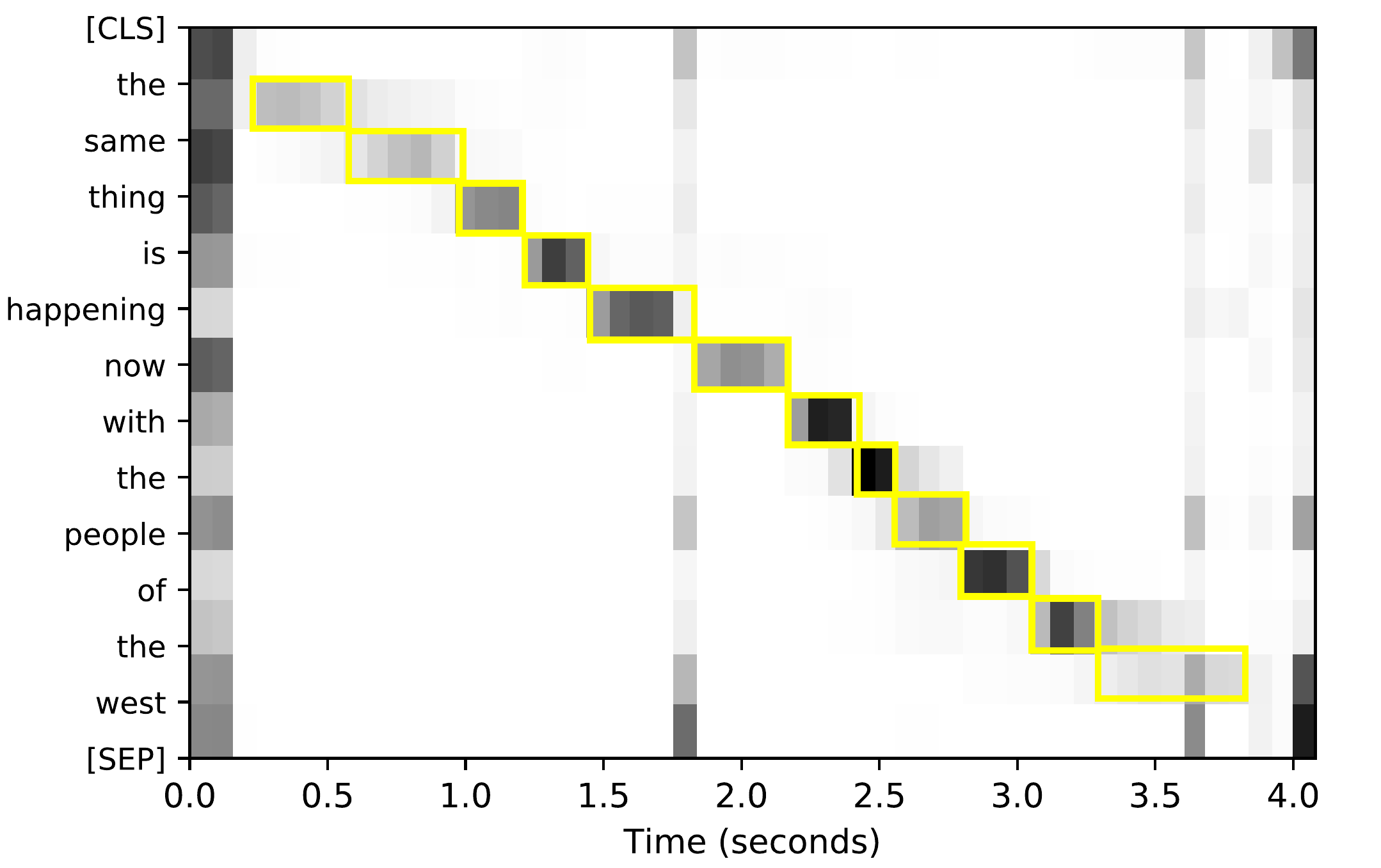}}
\caption{Attention heatmap of the cross-modal attention layer. The yellow boxes represent the actual alignment of the tokens with the speech. Cross-modal attention is implicitly learning alignment structures.}\vspace*{-0.05in}
\label{fig:heatmap}
\end{figure}

\noindent \textbf{Ablation study:} The last four rows of Table \ref{tab:results} present an ablation study. The Sequence Contrastive baseline serves as a coarse-grained counterpart of our proposed method where a sequence-level representation of a speech utterance is aligned with the corresponding representation from BERT. Tokenwise Contrastive pretaining significantly outperforms this baseline. This shows that the proposed pretraining is successful in learning a detailed alignment with BERT embeddings. SpecAugment significantly adds to the performance of our model especially in noisy acoustic settings and when speech data is limited.\\


\noindent \textbf{Cross-Modal attention analysis:} We used a cross-modal attention mechanism to learn an alignment between text tokens and speech frames in a completely unsupervised way. To see if such an alignment is being learnt by the cross-modal attention layer, we analyse the cross-modal attention heatmap from this layer for an utterance from the LibriSpeech dev-other set. This is shown in Figure \ref{fig:heatmap}. Note that there is a visible monotonic alignment along the diagonal of this heatmap which is very similar to the actual alignment of the tokens with the corresponding speech regions (shown in yellow). This shows that the model is successful in learning a tokenwise alignment between speech and text.

Also, note that a few regions in the speech are always attended to, irrespective of the token. For example, the first two speech frames always have a high attention weight. We hypothesize that these regions correspond to a location in the speech where the model embeds contextual information. Hence, the representation of a token can be seen as combination of the token-specific diagonal element and the context-specific fixed regions in the speech. A more detailed analysis of how the attention is being learnt is a subject for future work.

\section{Conclusion}

In this work, we proposed a new method for using pretrained BERT as a teacher to inject fine-grained token-level embedding information into speech representations. The proposed contrastive learning objective doesn't just learn speech-to-BERT alignment at the sentence level but rather at the WordPiece token level. We performed extensive experiments on two widely used S2I datasets and show that our proposed model reaches SOTA performance on both. For future work, it would be useful to look into ways of performing entity extraction and slot-filling using a model pretrained in the proposed way.

\section{Acknowledgement}
This work was partially supported by the National Science Foundation under Grant No. 2008043.



\bibliographystyle{IEEEtran}

\bibliography{mybib}

\begin{thebibliography}{10}
\providecommand{\url}[1]{#1}
\csname url@samestyle\endcsname
\providecommand{\newblock}{\relax}
\providecommand{\bibinfo}[2]{#2}
\providecommand{\BIBentrySTDinterwordspacing}{\spaceskip=0pt\relax}
\providecommand{\BIBentryALTinterwordstretchfactor}{4}
\providecommand{\BIBentryALTinterwordspacing}{\spaceskip=\fontdimen2\font plus
\BIBentryALTinterwordstretchfactor\fontdimen3\font minus
  \fontdimen4\font\relax}
\providecommand{\BIBforeignlanguage}[2]{{%
\expandafter\ifx\csname l@#1\endcsname\relax
\typeout{** WARNING: IEEEtran.bst: No hyphenation pattern has been}%
\typeout{** loaded for the language `#1'. Using the pattern for}%
\typeout{** the default language instead.}%
\else
\language=\csname l@#1\endcsname
\fi
#2}}
\providecommand{\BIBdecl}{\relax}
\BIBdecl

\bibitem{faruqui2021revisiting}
M.~Faruqui and D.~Hakkani-T{\"u}r, ``Revisiting the boundary between asr and
  nlu in the age of conversational dialog systems,'' \emph{Computational
  Linguistics}, pp. 1--12, 2021.

\bibitem{hsu2021hubert}
W.-N. Hsu, B.~Bolte, Y.-H.~H. Tsai, K.~Lakhotia, R.~Salakhutdinov, and
  A.~Mohamed, ``Hubert: Self-supervised speech representation learning by
  masked prediction of hidden units,'' \emph{IEEE/ACM Transactions on Audio,
  Speech, and Language Processing}, vol.~29, pp. 3451--3460, 2021.

\bibitem{baevski2020wav2vec}
A.~Baevski, Y.~Zhou, A.~Mohamed, and M.~Auli, ``wav2vec 2.0: A framework for
  self-supervised learning of speech representations,'' \emph{Advances in
  Neural Information Processing Systems}, vol.~33, pp. 12\,449--12\,460, 2020.

\bibitem{liu2019roberta}
Y.~Liu, M.~Ott, N.~Goyal, J.~Du, M.~Joshi, D.~Chen, O.~Levy, M.~Lewis,
  L.~Zettlemoyer, and V.~Stoyanov, ``Roberta: A robustly optimized bert
  pretraining approach,'' \emph{arXiv preprint arXiv:1907.11692}, 2019.

\bibitem{brown2020language}
T.~Brown, B.~Mann, N.~Ryder, M.~Subbiah, J.~D. Kaplan, P.~Dhariwal,
  A.~Neelakantan, P.~Shyam, G.~Sastry, A.~Askell \emph{et~al.}, ``Language
  models are few-shot learners,'' \emph{Advances in neural information
  processing systems}, vol.~33, pp. 1877--1901, 2020.

\bibitem{gopalakrishnan2020neural}
K.~Gopalakrishnan, B.~Hedayatnia, L.~Wang, Y.~Liu, and D.~Hakkani-T{\"u}r,
  ``Are neural open-domain dialog systems robust to speech recognition errors
  in the dialog history? an empirical study,'' in \emph{Proc. Interspeech},
  2020.

\bibitem{kim2021robust}
S.~Kim, Y.~Liu, D.~Jin, A.~Papangelis, K.~Gopalakrishnan, B.~Hedayatnia, and
  D.~Hakkani-Tur, ``" how robust ru?": Evaluating task-oriented dialogue
  systems on spoken conversations,'' \emph{arXiv preprint arXiv:2109.13489},
  2021.

\bibitem{stiff2019improving}
A.~Stiff, P.~Serai, and E.~Fosler-Lussier, ``Improving human-computer
  interaction in low-resource settings with text-to-phonetic data
  augmentation,'' in \emph{ICASSP 2019-2019 IEEE International Conference on
  Acoustics, Speech and Signal Processing (ICASSP)}.\hskip 1em plus 0.5em minus
  0.4em\relax IEEE, 2019, pp. 7320--7324.

\bibitem{serai2022hallucination}
P.~Serai, V.~Sunder, and E.~Fosler-Lussier, ``Hallucination of speech
  recognition errors with sequence to sequence learning,'' \emph{IEEE/ACM
  Transactions on Audio, Speech, and Language Processing}, 2022.

\bibitem{namazifar2021warped}
M.~Namazifar, G.~Tur, and D.~Hakkani-T{\"u}r, ``Warped language models for
  noise robust language understanding,'' in \emph{2021 IEEE Spoken Language
  Technology Workshop (SLT)}.\hskip 1em plus 0.5em minus 0.4em\relax IEEE,
  2021, pp. 981--988.

\bibitem{devlin2019bert}
J.~Devlin, M.-W. Chang, K.~Lee, and K.~Toutanova, ``Bert: Pre-training of deep
  bidirectional transformers for language understanding,'' in \emph{Proceedings
  of the 2019 Conference of the North American Chapter of the Association for
  Computational Linguistics: Human Language Technologies, Volume 1 (Long and
  Short Papers)}, 2019, pp. 4171--4186.

\bibitem{lugosch2019speech}
L.~Lugosch, M.~Ravanelli, P.~Ignoto, V.~S. Tomar, and Y.~Bengio, ``Speech model
  pre-training for end-to-end spoken language understanding,'' in \emph{Proc.
  Interspeech}, 2019.

\bibitem{ghannay2018end}
S.~Ghannay, A.~Caubri{\`e}re, Y.~Est{\`e}ve, N.~Camelin, E.~Simonnet,
  A.~Laurent, and E.~Morin, ``End-to-end named entity and semantic concept
  extraction from speech,'' in \emph{2018 IEEE Spoken Language Technology
  Workshop (SLT)}.\hskip 1em plus 0.5em minus 0.4em\relax IEEE, 2018, pp.
  692--699.

\bibitem{kuo2020end}
H.-K.~J. Kuo, Z.~T{\"u}ske, S.~Thomas, Y.~Huang, K.~Audhkhasi, B.~Kingsbury,
  G.~Kurata, Z.~Kons, R.~Hoory, and L.~Lastras, ``End-to-end spoken language
  understanding without full transcripts,'' in \emph{Proc. Interspeech}, 2020.

\bibitem{huang2020leveraging}
Y.~Huang, H.-K. Kuo, S.~Thomas, Z.~Kons, K.~Audhkhasi, B.~Kingsbury, R.~Hoory,
  and M.~Picheny, ``Leveraging unpaired text data for training end-to-end
  speech-to-intent systems,'' in \emph{ICASSP 2020-2020 IEEE International
  Conference on Acoustics, Speech and Signal Processing (ICASSP)}.\hskip 1em
  plus 0.5em minus 0.4em\relax IEEE, 2020, pp. 7984--7988.

\bibitem{kim2021two}
S.~Kim, G.~Kim, S.~Shin, and S.~Lee, ``Two-stage textual knowledge distillation
  for end-to-end spoken language understanding,'' in \emph{ICASSP 2021-2021
  IEEE International Conference on Acoustics, Speech and Signal Processing
  (ICASSP)}.\hskip 1em plus 0.5em minus 0.4em\relax IEEE, 2021, pp. 7463--7467.

\bibitem{kim2021st}
M.~Kim, G.~Kim, S.-W. Lee, and J.-W. Ha, ``St-bert: Cross-modal language model
  pre-training for end-to-end spoken language understanding,'' in \emph{ICASSP
  2021-2021 IEEE International Conference on Acoustics, Speech and Signal
  Processing (ICASSP)}.\hskip 1em plus 0.5em minus 0.4em\relax IEEE, 2021, pp.
  7478--7482.

\bibitem{rongali2021exploring}
S.~Rongali, B.~Liu, L.~Cai, K.~Arkoudas, C.~Su, and W.~Hamza, ``Exploring
  transfer learning for end-to-end spoken language understanding,'' in
  \emph{Proceedings of the AAAI Conference on Artificial Intelligence},
  vol.~35, no.~15, 2021, pp. 13\,754--13\,761.

\bibitem{chung2021splat}
Y.-A. Chung, C.~Zhu, and M.~Zeng, ``Splat: Speech-language joint pre-training
  for spoken language understanding,'' in \emph{Proceedings of the 2021
  Conference of the North American Chapter of the Association for Computational
  Linguistics: Human Language Technologies}, 2021, pp. 1897--1907.

\bibitem{qian2021speech}
Y.~Qian, X.~Bianv, Y.~Shi, N.~Kanda, L.~Shen, Z.~Xiao, and M.~Zeng,
  ``Speech-language pre-training for end-to-end spoken language
  understanding,'' in \emph{ICASSP 2021-2021 IEEE International Conference on
  Acoustics, Speech and Signal Processing (ICASSP)}.\hskip 1em plus 0.5em minus
  0.4em\relax IEEE, 2021, pp. 7458--7462.

\bibitem{chen2020simple}
T.~Chen, S.~Kornblith, M.~Norouzi, and G.~Hinton, ``A simple framework for
  contrastive learning of visual representations,'' in \emph{International
  conference on machine learning}.\hskip 1em plus 0.5em minus 0.4em\relax PMLR,
  2020, pp. 1597--1607.

\bibitem{chan2016listen}
W.~Chan, N.~Jaitly, Q.~Le, and O.~Vinyals, ``Listen, attend and spell: A neural
  network for large vocabulary conversational speech recognition,'' in
  \emph{2016 IEEE international conference on acoustics, speech and signal
  processing (ICASSP)}.\hskip 1em plus 0.5em minus 0.4em\relax IEEE, 2016, pp.
  4960--4964.

\bibitem{tuske2020single}
Z.~T{\"u}ske, G.~Saon, K.~Audhkhasi, and B.~Kingsbury, ``Single headed
  attention based sequence-to-sequence model for state-of-the-art results on
  switchboard,'' in \emph{Proc. Interspeech}, 2020.

\bibitem{vaswani2017attention}
A.~Vaswani, N.~Shazeer, N.~Parmar, J.~Uszkoreit, L.~Jones, A.~N. Gomez,
  {\L}.~Kaiser, and I.~Polosukhin, ``Attention is all you need,''
  \emph{Advances in neural information processing systems}, vol.~30, 2017.

\bibitem{panayotov2015librispeech}
V.~Panayotov, G.~Chen, D.~Povey, and S.~Khudanpur, ``Librispeech: an asr corpus
  based on public domain audio books,'' in \emph{2015 IEEE international
  conference on acoustics, speech and signal processing (ICASSP)}.\hskip 1em
  plus 0.5em minus 0.4em\relax IEEE, 2015, pp. 5206--5210.

\bibitem{park2019specaugment}
D.~S. Park, W.~Chan, Y.~Zhang, C.-C. Chiu, B.~Zoph, E.~D. Cubuk, and Q.~V. Le,
  ``Specaugment: A simple data augmentation method for automatic speech
  recognition,'' in \emph{Proc. Interspeech}, 2019.

\bibitem{saade2019spoken}
A.~Saade, J.~Dureau, D.~Leroy, F.~Caltagirone, A.~Coucke, A.~Ball, C.~Doumouro,
  T.~Lavril, A.~Caulier, T.~Bluche \emph{et~al.}, ``Spoken language
  understanding on the edge,'' in \emph{2019 Fifth Workshop on Energy Efficient
  Machine Learning and Cognitive Computing-NeurIPS Edition (EMC2-NIPS)}.\hskip
  1em plus 0.5em minus 0.4em\relax IEEE, 2019, pp. 57--61.

\bibitem{cha2021speak}
S.~Cha, W.~Hou, H.~Jung, M.~Phung, M.~Picheny, H.-K. Kuo, S.~Thomas, and
  E.~Morais, ``Speak or chat with me: End-to-end spoken language understanding
  system with flexible inputs,'' \emph{arXiv preprint arXiv:2104.05752}, 2021.

\bibitem{agrawal2020tie}
B.~Agrawal, M.~M{\"u}ller, M.~Radfar, S.~Choudhary, A.~Mouchtaris, and
  S.~Kunzmann, ``Tie your embeddings down: Cross-modal latent spaces for
  end-to-end spoken language understanding,'' \emph{arXiv preprint
  arXiv:2011.09044}, 2020.

\end{thebibliography}


\end{document}